\documentclass[10pt,twocolumn,letterpaper]{article}

\def\BUILDMODE{arxiv}

\usepackage{etoolbox}
\newtoggle{anon}
\ifdefstring{\BUILDMODE}{submission}{\toggletrue{anon}}{\togglefalse{anon}}

\ifdefstring{\BUILDMODE}{submission}{\usepackage[review,applications]{wacv}}{}
\ifdefstring{\BUILDMODE}{arxiv}{\usepackage[pagenumbers]{wacv}}{}
\ifdefstring{\BUILDMODE}{final}{\usepackage{wacv}}{}

\usepackage[utf8]{inputenc}
\usepackage[T1]{fontenc}
\usepackage{microtype}
\usepackage{adjustbox}   
\usepackage{stfloats}

\graphicspath{{figures/}{figures_static/}}

\newcommand{\emojiicon}[1]{\raisebox{-0.25ex}{\includegraphics[height=1.6ex]{#1}}}
\newcommand{\emojifire}{\emojiicon{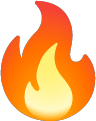}}
\newcommand{\emojisnowflake}{\emojiicon{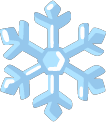}}

\definecolor{oodcolor}{HTML}{1E7A6E}
\newcommand{\ood}[1]{\textcolor{oodcolor}{#1}}

\newcommand{\anon}[2]{\iftoggle{anon}{#2}{#1}}

\newcommand{\repourl}{\anon{\url{https://github.com/JSeytre/chessqueries}}{}}
\newcommand{\demourl}{\anon{\url{https://chessq.org}}{}}

\newtoggle{showllm}
\toggletrue{showllm}

\definecolor{wacvblue}{rgb}{0.21,0.49,0.74}
\iftoggle{anon}{\PassOptionsToPackage{pagebackref}{hyperref}}{}
\usepackage[breaklinks,colorlinks,allcolors=wacvblue]{hyperref}
\def\wacvPaperID{179}
\def\confName{WACV}
\def\confYear{2027}

\title{ChessQueries: Toward Better Chess Board Recognition}

\author{Jo\"el Seytre\\
\texttt{contact@chessq.org}}

\iftoggle{anon}{\hypersetup{pdfauthor={},pdfsubject={},pdfkeywords={}}}{%
  \hypersetup{pdftitle={ChessQueries: Toward Better Chess Board Recognition},
              pdfauthor={Jo\"el Seytre},pdflang={en-US}}%
}

\begin{document}

%
\twocolumn[{%
  \renewcommand\twocolumn[1][]{##1}%
  \maketitle
  \begin{center}
    \setlength{\abovecaptionskip}{4pt}%
{\includegraphics[width=0.94\textwidth]{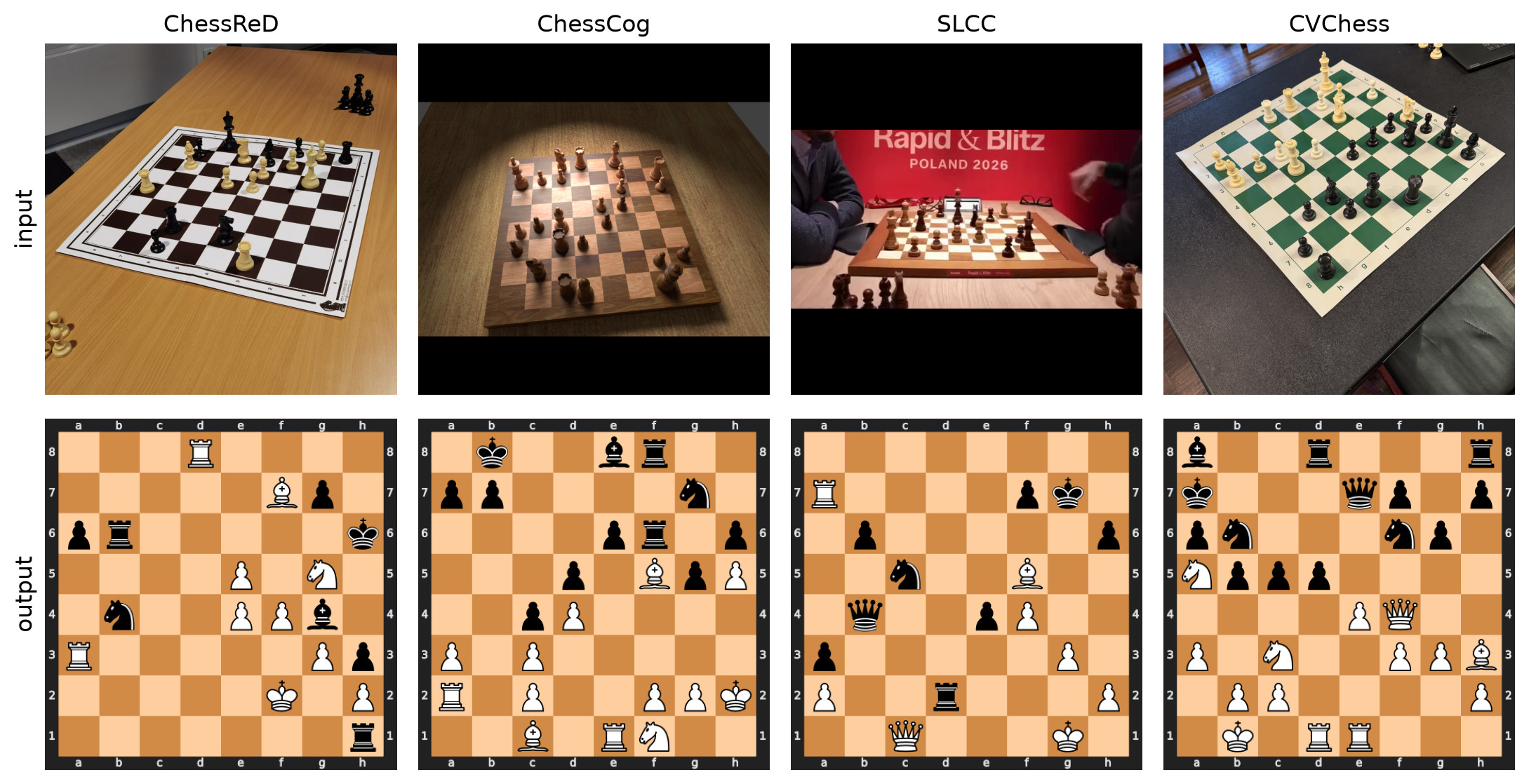}}
\captionof{figure}{ChessQueries can handle four datasets (ChessReD~\cite{chessred}, ChessCog~\cite{chesscog}, SLCC [ours], CVChess~\cite{abeykoon2025cvchess}) with various challenges.}
\label{fig:hero}

  \end{center}
}]

\begin{abstract}
%
Chess board recognition is the task of mapping the image of a chess board to the information of which piece is on which square. 
So far this task has two established benchmarks: ChessCog is synthetic, and ChessReD comes from smartphone pictures of a single chess board setup.
We introduce ChessQueries, a new method combining a ViT encoder with a DETR-style decoder, which outperforms existing methods.
On the ChessReD benchmark, we improve the state of the art from 15.3\% to 99.2\%, and demonstrate strong capabilities on out-of-distribution datasets.
Our method saturates the task on the two datasets, with an average 0.01 wrong squares per board (vs. SotA: 3.4 / 0.15 respectively).
We also share a new, harder public dataset, parsed from broadcasted top-level chess tournaments.
\footnote{
    \anon{Code, model weights and the SLCC data release: \repourl. Demo: \demourl.}
    {Code, model weights and the SLCC data will be released.}
}

\end{abstract}


\begin{figure*}[t]

  \centering

  \def\svgwidth{\textwidth}

  \includegraphics[width=\linewidth]{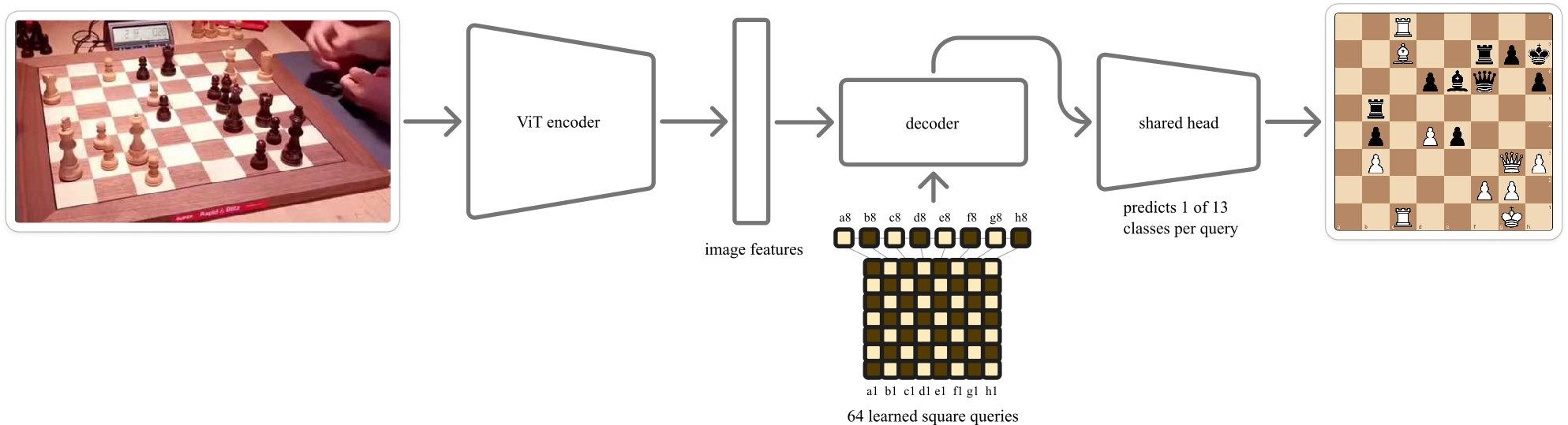}

  \caption{ChessQueries architecture. A ViT-L encoder turns the input image into patch tokens.
    Sixty-four learned \emph{square queries} --- one per board square ---
    cross-attend to those tokens, and a single shared linear head maps each
    decoded query to one of 13 classes (six piece types, times two colors, plus
    empty square), yielding the full $8\times8$ board in one forward pass.}

  \label{fig:arch}

\end{figure*}

\section{Introduction}
\label{sec:intro}

Chess board recognition consists of mapping an image of a board to its per-square state, such as the Forsyth-Edwards Notation (FEN) standard text format, which is used widely in the chess world.
This task has valuable applications for amateur play and analysis, and could be used to improve broadcasting of chess tournaments, 
during which technical difficulties from electronic sensory DGT \cite{DGT, DGT-patent} boards are common~\cite{fide2024technicalcommission, doggers2009dgt}.
It also represents a computer vision challenge, as can be seen in~\cref{fig:hero}, due to the varying camera angles, lighting, shadows, as well as partial occlusion.

The two main datasets used to date were either synthetic (ChessCog \cite{chesscog}) or created from smartphone pictures of games being played on a single physical chess board (ChessReD \cite{chessred}).
While methods for this task tend to focus on one of those datasets, we sought to establish a \emph{single} architecture that would work across datasets, 
but also in real chess broadcast conditions.

Our contributions are as follows: 
(1) we introduce a model called ChessQueries, which pairs a ViT encoder with a DETR-style decoder processing learned square queries, outperforming all existing methods, nearly saturating the existing test sets, with real generalization strengths;
(2) we demonstrate that our approach generalizes well to unseen datasets such as the single-game CVChess \cite{abeykoon2025cvchess} dataset, and is well-suited for training light-weight LoRAs on new domains;
(3) we introduce a new, hard 2,174-image chess board recognition dataset based on the Saint Louis Chess Club YouTube broadcasts \cite{stlcc, lichess, GCT}, with a new unique level of challenge (lighting, partial occlusions, hard viewpoints).

\begin{table*}[t]
\centering
\adjustbox{max width=\linewidth}{%
\begin{tabular}{l@{\hskip 1.2em}cc@{\hskip 1.2em}cc@{\hskip 1.2em}cc@{\hskip 1.2em}cc}
\toprule
 & \multicolumn{2}{c}{\shortstack{ChessReD\\[1pt] $n{=}2129$}} & \multicolumn{2}{c}{\shortstack{ChessCog\\[1pt] $n{=}342$}} & \multicolumn{2}{c}{\shortstack{SLCC\\[1pt] $n{=}373$}} & \multicolumn{2}{c}{\shortstack{CVChess\\[1pt] $n{=}352$}} \\
\cmidrule(lr){2-3}\cmidrule(lr){4-5}\cmidrule(lr){6-7}\cmidrule(lr){8-9}
Method & Board\,$\uparrow$ & Wrong sq.\,$\downarrow$ & Board\,$\uparrow$ & Wrong sq.\,$\downarrow$ & Board\,$\uparrow$ & Wrong sq.\,$\downarrow$ & Board\,$\uparrow$ & Wrong sq.\,$\downarrow$ \\
\midrule
\multicolumn{9}{l}{\emph{Trained on ChessCog\,+\,ChessReD\,+\,SLCC}} \\
\quad ChessReD, original recipe & 11.5\% & 5.0 & 0\% & 21.1 & 0\% & 20.1 & \ood{0\%} & \ood{37.1} \\
\quad ChessReD, generalizing recipe & 40.3\% & 1.3 & 60.5\% & 0.68 & 26.0\% & 2.4 & \ood{8.8\%} & \ood{7.2} \\
\quad ChessQueries [ours] & \textbf{99.5\%} & \textbf{0.01} & \textbf{98.5\%} & \textbf{0.01} & \textbf{87.1\%} & \textbf{0.25} & \ood{\textbf{87.6\%}} & \ood{\textbf{0.50}} \\
\midrule
\multicolumn{9}{l}{\emph{Trained on ChessReD only}} \\
\quad ChessReD (published)~\cite{chessred} & 15.3\% & 3.4 & \ood{0\%} & \ood{47.9} & \ood{0\%} & \ood{44.3} & \ood{0\%} & \ood{54.6} \\
\quad ChessReD, generalizing recipe & 12.2\% & 4.4 & \ood{0\%} & \ood{42.9} & \ood{0\%} & \ood{30.8} & \ood{0.57\%} & \ood{11.5} \\
\quad ChessQueries [ours] & \textbf{99.2\%} & \textbf{0.01} & \ood{\textbf{0.29\%}} & \ood{\textbf{13.6}} & \ood{\textbf{0.14\%}} & \ood{\textbf{16.5}} & \ood{\textbf{56.0\%}} & \ood{\textbf{2.9}} \\
\midrule
\multicolumn{9}{l}{\emph{Trained on ChessCog only}} \\
\quad ChessCog~\cite{chesscog} & \textbf{2.3\%}\textsuperscript{\S} & 42.9\textsuperscript{\S} & 93.9\% & 0.15 & \ood{\textbf{0\%}}\textsuperscript{\S} & \ood{34.0}\textsuperscript{\S} & \ood{\textbf{0\%}}\textsuperscript{\S} & \ood{\textbf{20.8}}\textsuperscript{\S} \\
\quad ChessReD (published)~\cite{chessred} & -- & -- & 39.8\% & 1.2 & -- & -- & -- & -- \\
\quad ChessQueries [ours] & \ood{0.54\%} & \ood{\textbf{21.7}} & \textbf{98.2\%} & \textbf{0.02} & \ood{\textbf{0\%}} & \ood{\textbf{29.6}} & \ood{\textbf{0\%}} & \ood{24.8} \\
\midrule
\multicolumn{9}{l}{\emph{Frontier LLM baseline}} \\
\quad Claude Opus 5\textsuperscript{\dag}~\cite{anthropic2026claudeopus5} & \textbf{1.5\%}\textsuperscript{\ddag} & \textbf{21.6}\textsuperscript{\ddag} & 0\% & \textbf{21.7} & \textbf{0\%} & \textbf{23.0} & \textbf{2.6\%} & \textbf{18.9} \\
\quad GPT-5.6 Sol\textsuperscript{\dag}~\cite{openai2026gpt56sol} & 0.50\%\textsuperscript{\ddag} & 24.9\textsuperscript{\ddag} & \textbf{0.29\%} & 23.9 & \textbf{0\%} & 25.7 & 0.57\% & 24.9 \\
\bottomrule
\end{tabular}
}
\caption{Overall performance. We report \emph{exact-board accuracy} and the \emph{average number of wrong squares} (out of 64) on the test sets. CVChess is purely a test set (only 1 game). ChessQueries achieves the best results across categories, and generalizes the best on unseen domains (\ood{highlighted in teal}); top individual performances observed when training on all domains. \textsuperscript{\S}ChessCog board corner detection fails on 65.7\% / 96.8\% / 50.0\% of ChessReD / SLCC / CVChess respectively, so the reported wrong squares numbers are averaged over the correctly localized boards. ChessReD performance is reported by~\cite{chessred}, as they followed ChessCog's protocol and fine-tuned it on two ChessReD starting-position images and took the best results of all possible board orientations. SLCC and CVChess numbers are our own zero-shot runs of their released pipeline.\textsuperscript{\dag}For multi-modal LLMs: images square-resized to $644{\times}644$ (to match regular input conditions), 32k max output tokens, structured JSON output; Claude Opus 5 at its minimal \texttt{effort{=}low}, GPT-5.6 at \texttt{effort{=}none}. On a subset of 40 images, increased \texttt{effort} slightly improved per-square accuracy but per-board accuracy remained at 0\%, so lowest effort was chosen to save on costs. A few-shot approach with example input/output from the training sets had no impact. 0.5--2\% of outputs were invalid FEN chess positions, so the overall frontier LLM failure is indeed a visual understanding one. In a separate experiment, we observed that Qwen3-VL-8B~\cite{bai2025qwen3vl} fails the same way ($0.5\%$ board accuracy on ChessReD), yet a LoRA fine-tune of that same model reaches $87\%$: what the frontier models lack here is task-specific visual training. \textsuperscript{\ddag} Using a subset of 400 / 2129 ChessReD test images to limit costs.}
\label{tab:main}
\end{table*}

\section{Related Work}
\label{sec:related}

\paragraph{Multi-stage pipelines.}
A traditional approach to chess recognition is to leverage a multi-stage pipeline.
It decomposes recognition into board detection, square localization, and per-square piece classification~\cite{neufeld2010probabilistic,xie2018geometry,xie2018chesspiece,
czyzewski2020chessboard,mallasen2020livechess2fen}.
Specifically, \emph{chesscog}~\cite{chesscog} fits the board with a RANSAC-based projective transform 
and then runs separate CNNs for occupancy and piece prediction; the method comes with a synthetic, 
Blender-rendered dataset of $4{,}888$ images, an idea already explored in~\cite{neto2019chessposition}.
The method requires knowing from which player's perspective (white or black) 
the image was taken.  
CVChess~\cite{abeykoon2025cvchess} follows a similar approach, with Hough-line board detection, 
projective warp to a top-down view, broken down into 64 squares, with an eventual CNN 
mapping each square crop to one of 13 states (six white pieces, six black, empty).  
Such pipelines are accurate when every successive stage succeeds, but errors compound: 
it has been established that ChessCog's detector, tuned on synthetic imagery, 
localizes the real-life boards of ChessReD's real photographs~\cite{chessred} only $34.4\%$ of the time.
Both systems depend on explicit geometric cues (e.g., detected corners, supplied orientation) 
that are unreliable on unconstrained or out-of-domain images.

\paragraph{End-to-end recognition.}
This motivated the authors of ChessReD~\cite{chessred} to remove the intermediate 
stages, predicting the full board configuration directly from the image, in an end-to-end fashion. 
They released ChessReD: $10{,}800$ real smartphone photographs of 100 games across 
three cameras with varied angles and lighting. This constitutes the first large-scale \emph{non-synthetic}
benchmark, split game-wise to avoid leakage between train and test sets.
Their model used a ResNeXt-101~\cite{xie2017resnext} classifier, along with a prediction head 
for each of the 64 squares' 13 possible outcomes.
Interestingly, they also tried a set-prediction variant, drawing inspiration from DETR \cite{carion2020detr} 
to predict chess row and file coordinates of the pieces on the board. 
They reported that it failed to converge, and attributed this to the difficulty of small pieces. 
In this work, we will show that we also draw inspiration from DETR,
but in a different way that focuses on the \emph{object queries} introduced in the original paper.
We draw inspiration from prior work on leveraging learned vectors 
for structured outputs~\cite{lee2019settransformer,jaegle2022perceiverio, locatello2020slotattention}.  

\paragraph{Cross-domain generalization.}
Each original method described above was developed for a single domain, and the
cross-domain experiments reported by ChessReD~\cite{chessred} are disappointing.
ChessReD's authors report that, after following ChessCog's protocol to adapt the model to new chess pieces, 
ChessCog reached only $2\%$ board accuracy on ChessReD (vs. their $15\%$).
Conversely, when they trained their ResNeXt architecture on ChessCog, it reached only $40\%$ (vs. their $94\%$). 
Neither method was able to be competitive out of its original domain.

In this work, we present a single model that can address the different domains of
synthetic renders and real-world photographs (the ChessReD dataset 
as well as the $352$ test images from the single-game released with CVChess~\cite{abeykoon2025cvchess}). 
We will also explore whether our model can easily be adapted to new, unseen domains.

\section{Method}
\label{sec:method}


We treat board recognition as structured per-square prediction over a fixed
domain model: a board is exactly 64 squares, each labelled with one of 13
classes (six white pieces, six black, or empty). \cref{fig:arch} shows the
architecture. 
We use a ViT-L/14 \cite{dosovitskiy2021vit} encoder ($304$M parameters), initialised from DINOv2 \cite{oquab2024dinov2}.
It maps the $644{\times}644$ input image to a set of patch tokens. 
The sixty-four \emph{square queries} are embeddings that are learned for each individual square, 
taking as input its square ID as well as its rank, file and color.  

The queries cross-attend to the image tokens through a 4-layer DETR-style decoder~\cite{carion2020detr},
and a single shared linear head maps each decoded query to its per-square class.  
Each square query is trained with a specific board square as its target, and as such there is no Hungarian matching needed.  

\begin{figure}[t]
  \centering
  \includegraphics[width=0.90\linewidth]{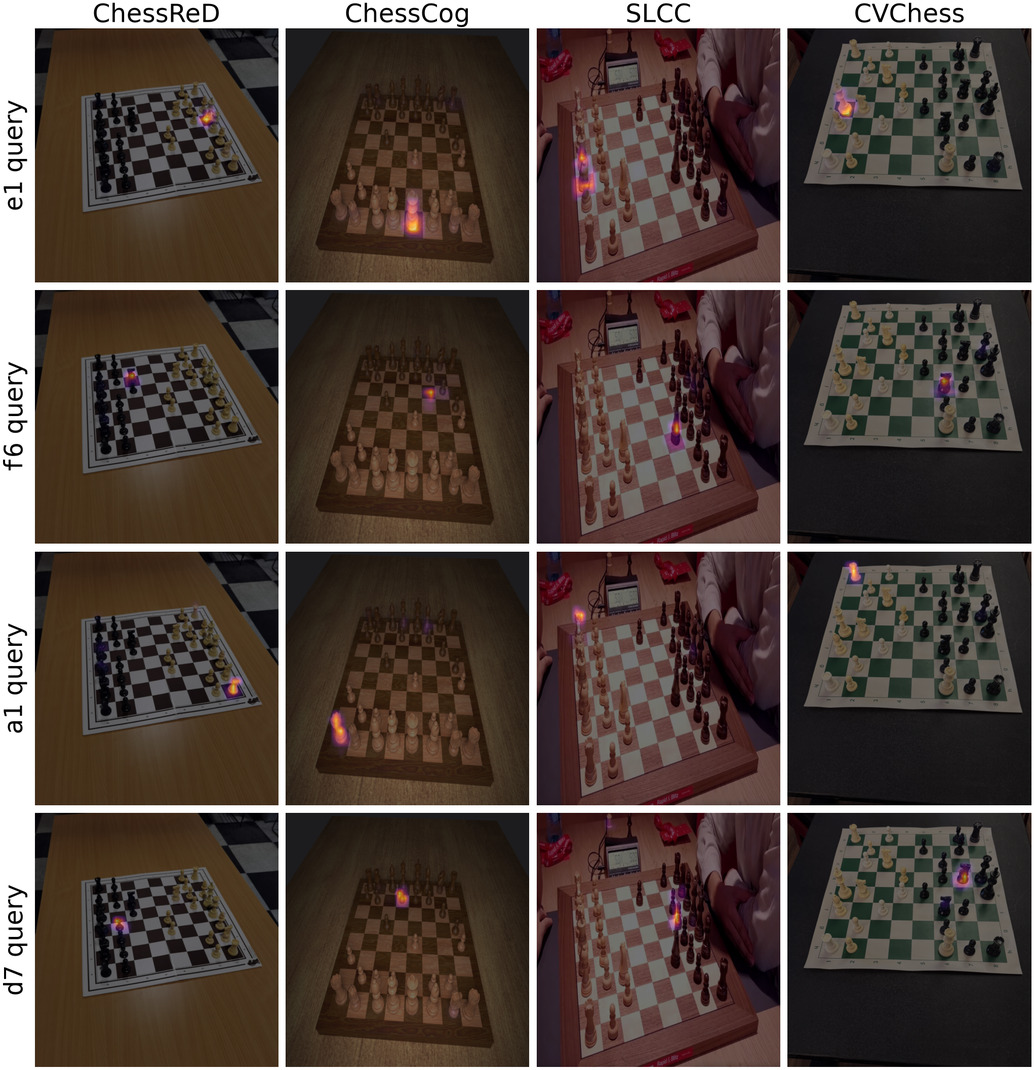}
  \caption{Query attention for 4 separate square queries.}
  \label{fig:attention}
\end{figure}

Examining the attention maps of each square query across datasets shows that 
the attention is trained to focus accurately on each respective square (see \cref{fig:attention}).  

The whole board is produced in one forward pass, with no required intermediate output 
such as board detection or corner estimation.  
Our approach also handles inputs of any orientation, whether seen from one of the players' 
perspectives, or from the side. 

\paragraph{Use of Large Language Models.}

\enlargethispage{1.5\baselineskip}

Large language models assisted in this work in two distinct roles. 
First, use of Claude Code (with various usage of Fable / Opus 4.8 / Opus 5 / Sonnet 5 / GPT 5.6 Sol) 
assisted in writing the code and running the experiments. 
The coding assistant developed plans, which we reviewed, 
were implemented by the assistant and then reviewed by us as a GitHub pull request, 
similarly to how a standard developer would contribute to a repository. 
Tests and other good coding practices were enforced to maintain high code quality, 
as can be seen in the code shared alongside this project.  

As for writing assistance, Claude Code was also used to structure and tweak the formatting in \LaTeX, 
the figures and tables, as well as grammar and spell-check, 
but the core of the content (structure, wording, messaging) was hand-written.
The authors take full responsibility for all content, including all findings, numbers, and citations.

\section{Experiments}
\label{sec:experiments}

\begin{figure*}[t]
  \centering
  \includegraphics[width=\linewidth]{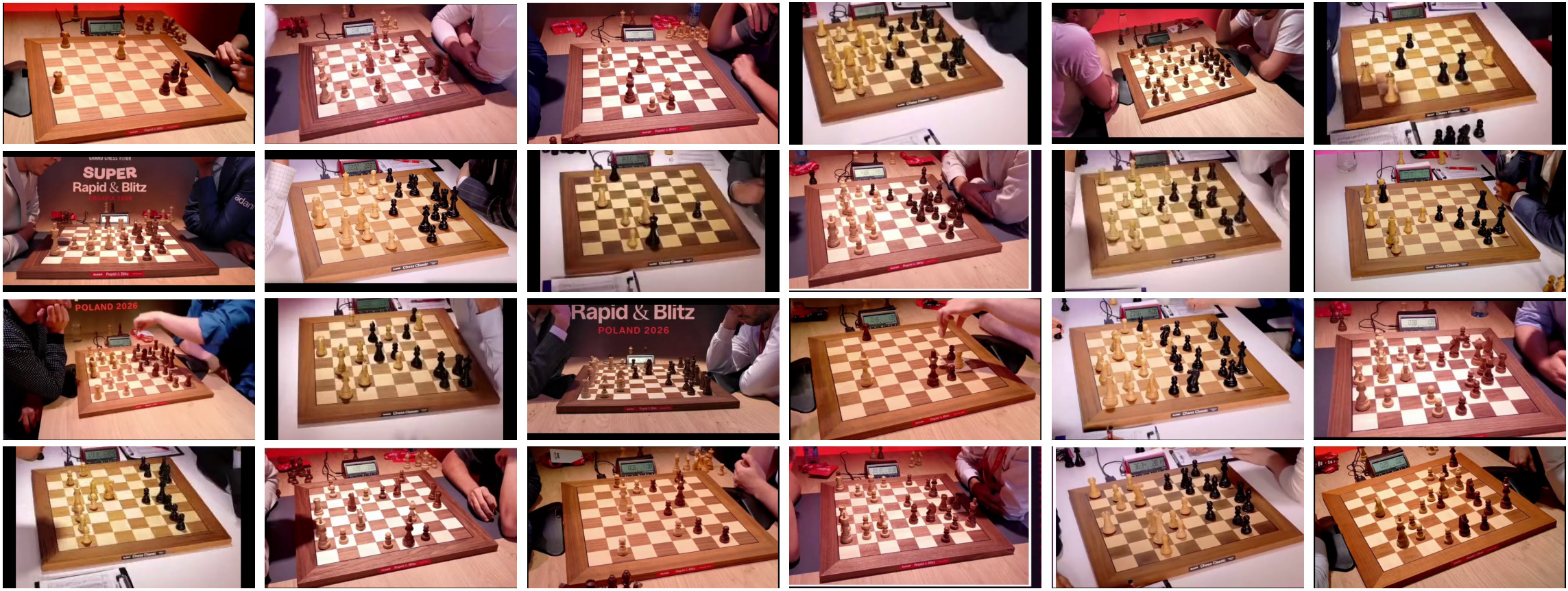}
  \caption{Samples from the SLCC dataset: 24 frames drawn from the 20 different broadcasts. The camera angle, lighting, board and piece set, background clutter and player/hand occlusions all vary, making the task challenging.}
  \label{fig:slcc_gallery}
\end{figure*}

\paragraph{Datasets.}
For this work we use 4 key datasets: 
(1) ChessReD~\cite{chessred} is a real-life dataset based on photographs of 100 games using a single chess board; 
(2) ChessCog~\cite{chesscog} consists of purely synthetic Blender renders; 
(3) CVChess~\cite{abeykoon2025cvchess} is similar to ChessReD, except that only a single game was recorded with multiple challenging camera angles of each position (thus we always use it as an out-of-domain test set);
and finally (4) SLCC (Saint Louis Chess Club), a broadcast dataset we created (see \cref{tab:data}).

\paragraph{The SLCC dataset.}
We built the 2,174 images of SLCC from 20 Saint Louis Chess Club~\cite{stlcc} / Grand Chess Tour~\cite{GCT} broadcast
videos, spanning three 2026 multi-day tournaments in Poland, Romania and Croatia: we parsed the YouTube videos 
automatically into templates whose layouts were hand-labeled, and verified all positions by seeking consensus from three sources:
(1) the Lichess~\cite{lichess} relay of the games, where we automatically identified the correct ply by parsing the player names and remaining clock times with OCR~\cite{du2020ppocr};
(2) a fine-tuned LoRA \cite{hu2021loralowrankadaptationlarge} of our model after having annotated the first 20 SLCC images; 
(3) a human review of the outputs, as every retained sample was human-verified.  

In practice, the model was used to rank multiple candidate relay positions from lichess (there often was a slight delay in the broadcast). 
No mistakes were found when the Lichess clock-time matching method and the model agreed.  
The images are particularly challenging due to partial obstruction, viewpoint, 
lighting and sometimes low resolution, due to only a small part of the broadcast showing the chess board.
That said, during the human review we made sure that no image was kept 
where the task was impossible for the model due to full occlusion of pieces or squares (e.g., by a player's hand).
\cref{fig:slcc_gallery} shows a sample of the resulting frames, and more details can be found in~\cref{sec:supp_slcc_annotation}.

SLCC is obtained from publicly available YouTube broadcasts, and the dataset and labeling code used \anon{are openly available}{will be made openly available}.
Following established annotation-based dataset releases based on
YouTube-sourced videos~\cite{kay2017kinetics,gemmeke2017audio,real2017youtube,gu2018ava,zhou2018youcook2,tang2019coin,chen2020vggsound},
we distribute video identifiers, timestamps, frame crop coordinates, extraction tooling, and chess position labels, but not the frames themselves.
Individuals are occasionally visible in the frames; these are public figures, i.e.,\ professional chess players appearing in publicly broadcast tournaments 
(e.g., Maxime Vachier-Lagrave and current world chess champion Gukesh Dommaraju in \cref{fig:slcc_worst}).

The release is licensed under CC BY-NC 4.0, for non-commercial research use only.

We have reached out to the SLCC to inform them of this work (we have received no reply to date).
For future work, our approach could be scaled to more chess broadcast videos, leveraging our shared method and code.


\paragraph{Our results.}
We trained on a single GeForce RTX 4090 for 45 epochs. We noticed that starting with a frozen encoder for the first 5 epochs improved training stability.
At inference, one forward pass reads the full board in \textbf{19\,ms} on the 4090 
($\sim$52 images/s in \texttt{bf16} at $644{\times}644$ resolution, with batch size one), 
i.e.,\ compatible with real-time live broadcasting use.

We trained with batch size $6$ in \texttt{bf16} mixed precision, optimizing a simple per-square $13$-way softmax cross-entropy loss 
(we tried adding auxiliary losses for piece colors and types, but they were not helpful).
We used AdamW~\cite{loshchilov2019adamw} (weight decay $0.05$, gradient clipping at $1.0$) with a cosine learning-rate schedule and a $3$-epoch linear warmup, 
peaking at $1.4{\times}10^{-4}$ for the decoder and readout and at $1.4{\times}10^{-5}$ for the encoder; 
the same warmup was re-applied to the encoder when it was unfrozen at epoch $5$.
Inputs were augmented with mild geometric transforms (random rotation up to $45^\circ$, perspective distortion, scaling in $[0.85, 1.1]$ and small translations) together with light color jitter, and we kept the checkpoint with the best validation exact-board accuracy.
Every number reported for our model in \cref{tab:main} is the mean over three independent seeds; the ablations in \cref{tab:ablations} and the head comparison in \cref{tab:head} used two seeds per configuration.

\begin{table}[t]
\centering
\begin{tabular}{lrrr}
\toprule
Split & Games & Shots & Frames \\
\midrule
Train & 106 & 768 & 1475 \\
Val & 23 & 128 & 326 \\
Test & 23 & 160 & 373 \\
\midrule
Total & 152 & 1056 & 2174 \\
\bottomrule
\end{tabular}
\caption{SLCC, the broadcast dataset we introduce: crops from 20 Saint Louis Chess Club / Grand Chess Tour broadcast videos on Youtube. Ground-truth positions are matched through the Lichess relay. \emph{Shots} counts the distinct camera shots that contributed at least one annotated frame. Splits are game-wise (no game spans two splits) to avoid evaluation leakage.}
\label{tab:data}
\end{table}

\cref{tab:main} is the headline result: our model advances the state of the art in every setting.
Our best results were from training on all training sets jointly, where we nearly saturated the ChessReD \& ChessCog benchmarks,
with 99.5\% / 98.5\% perfect board prediction, and 0.01 average wrong square per board (i.e.,\ 1 wrong square predicted every $\sim$6400).  

Under the same training data conditions as each corresponding published method, our model outperformed ChessReD and ChessCog as follows: 
on ChessReD, we achieved 99.2\% exact board accuracy, compared to their 15.3\%. 
On ChessCog, we achieved 98.2\%, compared to their 93.9\%.

Additionally, we reproduced the ChessReD model and trained two versions on the joint ChessReD + ChessCog + SLCC training set, 
one faithful to the original paper, and one recipe better suited for generalization and training on multiple datasets.
The "generalizing recipe" transplants our own training setup onto the unchanged ResNeXt-101 (32$\times$8d) architecture, 
changing five things relative to the original:
(1) binary cross-entropy on one-hot targets $\rightarrow$ per-square $13$-way softmax cross-entropy;
(2) Adam~\cite{kingma2015adam} at $10^{-3}$ with a $\times0.1$ step decay $\rightarrow$ AdamW ($1.4{\times}10^{-4}$, weight decay $0.05$) under a cosine schedule;
(3) $1024{\times}1024$ ChessReD-normalized inputs $\rightarrow$ $644{\times}644$ ImageNet-normalized inputs;
(4) no augmentation $\rightarrow$ the same geometric augmentation as our model, with gradient clipping at $1.0$;
(5) $200$ epochs $\rightarrow$ $45$ epochs.

This isolates the architecture as the single variable against our model.
We outperform the generalizing recipe significantly, 40.3\% $\rightarrow$ 99.5\% on ChessReD.

We also note the stark difference between the ChessReD and ChessCog data domain, 
as no model trained on one dataset performs well on the other.

\section{Analysis}
\label{sec:analysis}


\paragraph{Query attention.}
The learned square queries give a direct interpretability handle: each query's
cross-attention can be visualized (\cref{fig:attention}).
It localises to a single physical square and its piece, across domains and viewpoints. 
When the image features heavy occlusion, the attention focuses on the parts of the piece 
that appear in-between the surrounding pieces (see the e1 king from the SLCC sample).
Reading across any row, the same query stays on its square as the board's style, lighting, and
perspective change from real photographs to synthetic renders and broadcast stills.

\paragraph{Ablations.}
\cref{tab:ablations} introduces one modification at a time to input resolution, encoder scale, and augmentation. 
ChessReD and ChessCog stay near-saturated under every variant (within 3 board points of the main model), 
so the performance impact of these changes is observed on the harder domains, especially out-of-distribution domains. 
Encoder scale is the largest factor for zero-shot generalization: 
swapping ViT-L for ViT-B barely moves the in-domain numbers 
but reduces zero-shot CVChess from $87.6\%$ to $51.4\%$ board accuracy. 
Removing geometric augmentation is similarly costly ($41.2\%$ board accuracy,
$6.6$ wrong squares), consistent with CVChess's extreme camera angles.
Lowering the resolution from $644$ to $448$ pixels impacts the two near-saturated benchmarks much less than
the hard broadcast domain (SLCC $87.1\% \rightarrow 81.6\%$), where the board occupies a small, low-resolution part of the frame.
\begin{table*}[t]
\centering
\adjustbox{max width=\linewidth}{%
\begin{tabular}{lccc@{\hskip 1.2em}cc@{\hskip 1.2em}cc@{\hskip 1.2em}cc@{\hskip 1.2em}cc}
\toprule
 & \multicolumn{3}{c}{Recipe} & \multicolumn{2}{c}{ChessReD} & \multicolumn{2}{c}{ChessCog} & \multicolumn{2}{c}{SLCC} & \multicolumn{2}{c}{CVChess} \\
\cmidrule(lr){2-4}\cmidrule(lr){5-6}\cmidrule(lr){7-8}\cmidrule(lr){9-10}\cmidrule(lr){11-12}
Model & Res. & Backbone & Aug. & Board\,$\uparrow$ & Wrong sq.\,$\downarrow$ & Board\,$\uparrow$ & Wrong sq.\,$\downarrow$ & Board\,$\uparrow$ & Wrong sq.\,$\downarrow$ & Board\,$\uparrow$ & Wrong sq.\,$\downarrow$ \\
\midrule
ChessQueries (main) & 644 & ViT-L & \checkmark & \textbf{99.5\%} & \textbf{0.01} & \textbf{98.5\%} & \textbf{0.01} & \textbf{87.1\%} & \textbf{0.25} & \textbf{87.6\%} & \textbf{0.50} \\
Lower resolution & \textbf{448} & ViT-L & \checkmark & 98.4\% & 0.02 & 98.1\% & 0.02 & 81.6\% & 0.39 & 74.0\% & 1.8 \\
Smaller encoder & 644 & \textbf{ViT-B} & \checkmark & 99.2\% & 0.01 & 97.5\% & 0.02 & 84.2\% & 0.34 & 51.4\% & 3.9 \\
Smallest encoder & 644 & \textbf{ViT-S} & \checkmark & 98.6\% & 0.01 & 97.7\% & 0.02 & 75.9\% & 0.51 & 49.4\% & 1.2 \\
No augmentation & 644 & ViT-L & \textbf{$\times$} & 97.4\% & 0.03 & 95.9\% & 0.04 & 78.8\% & 0.48 & 41.2\% & 6.6 \\
\bottomrule
\end{tabular}
}
\caption{Ablation study: changed parameters are in bold. Contributions are mainly observed on the hard domains (SLCC, CVChess), notably the backbone size; augmentations have the biggest impact, most of all on CVChess as that dataset presents extreme view angles. ChessReD/ChessCog stay near-saturated throughout, with less than 3 percentage points regression.}
\label{tab:ablations}
\end{table*}

\begin{table*}[b]
\centering
\adjustbox{max width=\linewidth}{%
\begin{tabular}{l@{\hskip 1.2em}cc@{\hskip 1.2em}cc@{\hskip 1.2em}cc@{\hskip 1.2em}cc}
\toprule
 & \multicolumn{2}{c}{ChessReD} & \multicolumn{2}{c}{ChessCog} & \multicolumn{2}{c}{SLCC} & \multicolumn{2}{c}{CVChess} \\
\cmidrule(lr){2-3}\cmidrule(lr){4-5}\cmidrule(lr){6-7}\cmidrule(lr){8-9}
Method & Board\,$\uparrow$ & Wrong sq.\,$\downarrow$ & Board\,$\uparrow$ & Wrong sq.\,$\downarrow$ & Board\,$\uparrow$ & Wrong sq.\,$\downarrow$ & Board\,$\uparrow$ & Wrong sq.\,$\downarrow$ \\
\midrule
\multicolumn{9}{l}{\emph{\emojifire~Trained on ChessCog\,+\,ChessReD\,+\,SLCC}} \\
\quad Query decoder & \textbf{99.5\%} & \textbf{0.01} & \textbf{98.5\%} & \textbf{0.01} & \textbf{87.1\%} & \textbf{0.25} & \ood{\textbf{87.6\%}} & \ood{\textbf{0.50}} \\
\quad Linear head & 99.0\% & \textbf{0.01} & 98.4\% & 0.02 & 83.9\% & 0.30 & \ood{72.2\%} & \ood{2.1} \\
\midrule
\multicolumn{9}{l}{\emph{\emojifire~Trained on ChessCog\,+\,ChessReD}} \\
\quad Query decoder & \textbf{99.3\%} & \textbf{0.01} & \textbf{99.1\%} & \textbf{0.01} & \ood{\textbf{0.27\%}} & \ood{\textbf{16.2}} & \ood{\textbf{77.1\%}} & \ood{\textbf{1.2}} \\
\quad Linear head & 98.8\% & 0.03 & 98.7\% & \textbf{0.01} & \ood{0\%} & \ood{25.3} & \ood{38.6\%} & \ood{5.4} \\
\midrule
\multicolumn{9}{l}{\emph{\emojisnowflake~Frozen encoder, trained on ChessCog\,+\,ChessReD\,+\,SLCC}} \\
\quad Query decoder & \textbf{92.4\%} & \textbf{0.11} & \textbf{93.4\%} & \textbf{0.07} & \textbf{53.0\%} & \textbf{1.4} & \ood{\textbf{17.8\%}} & \ood{\textbf{7.1}} \\
\quad Linear head & 0\% & 20.2 & 0\% & 21.0 & 0\% & 19.9 & \ood{0\%} & \ood{23.6} \\
\bottomrule
\end{tabular}
}
\caption{Query decoder vs.\ linear head. The encoder is either fine-tuned (\emojifire) or frozen at its DINOv2 initialization (\emojisnowflake). Out-of-domain numbers are \ood{highlighted in teal}. At their best, the two heads are near-identical in-domain. The square query + decoder approach pulls ahead \ood{out of domain}. On a frozen encoder (last 2 rows), the contrast is stark: the query decoder still manages to learn the task, whereas the linear head remains stuck at 0\%. }
\label{tab:head}
\end{table*}

\paragraph{Query decoder vs.\ linear head.}
We seek to isolate the contribution of the square query + decoder head design
by replacing it with a simple $8{\times}8$ grid-pooling of the encoder features, leading into a plain linear head.  
We show the results in \cref{tab:head}, and the resulting architecture in the supplementary materials (\cref{fig:linhead_arch}).  
Trained on all three training sets, the two heads are near-identical in-domain, 
but we observe that the square query design is preferable for (1) out-of-domain performance,
(2) ease of training, and (3) better localization capabilities.

Indeed, we observe that a performance gap opens under domain shift (see \ood{teal numbers} in \cref{tab:head}): 
with SLCC held out of training, the decoder gets
$16.2$ wrong squares per board on SLCC against the linear head's $25.3$, and
reaches $77.1\%$ zero-shot board accuracy on CVChess against $38.6\%$. 
The query decoder is therefore better for generalization and, 
as we show next, it also comes with interpretability.

Additionally, we compared the two approaches after freezing the ViT encoder at its DINOv2 \cite{oquab2024dinov2} initialization, 
and we observed that the query decoder still manages to reach $92$--$94\%$ on ChessReD / ChessCog, and $53\%$ on SLCC,
whereas the linear head baseline is stuck at $0\%$ (predicting only the  majority class: empty squares), 
across all hyperparameters we tested.
We can thus conclude that the fine-tuning of the encoder reorganizes the encoded features 
into a grid that per-cell readout can consume, whereas the square query decoder can 
compute the visual square correspondence itself 
(see \cref{fig:frozen_attention} in the supplementary materials).
We observe that freezing the encoder negatively impacts our square query approach, as CVChess performance decreases from $87.6\%$ to $17.8\%$.

Finally, looking at the encoder attention maps, we can see that fine-tuning the ViT encoder with a linear head
results in less precise localization of the squares and their associated pieces 
(comparing \cref{fig:attention} and \cref{fig:linhead_attention}).

\begin{figure*}[t]
  \centering
  \includegraphics[width=0.88\linewidth]{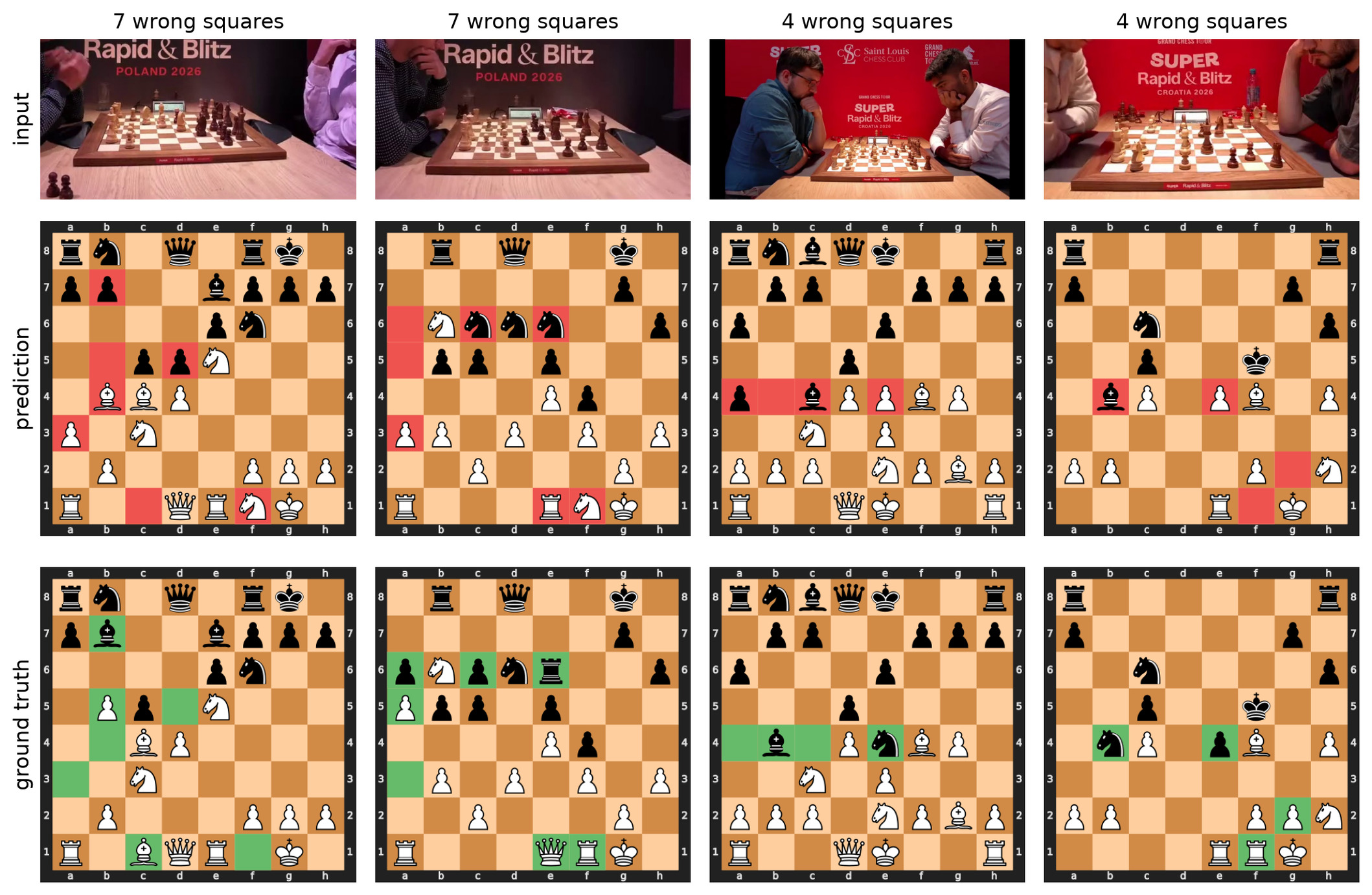}
  \caption{The 4 SLCC test images where ChessQueries performs worst. Errors happen under heavy occlusion, and remain difficult for humans.}
  \label{fig:slcc_worst}
\end{figure*}

\paragraph{Where the model still fails.}
Our worst SLCC test set predictions are shown in \cref{fig:slcc_worst} (up to seven squares mistaken out of 64).
The mistakes cluster on areas under heavy occlusion, where pieces are barely visible. 
The equivalent worst cases for the other three datasets are shown in the supplementary materials \cref{sec:supp}.

\paragraph{Few-shot experimentation protocol.}
We sought to explore few-shot experiments, to determine how many training images 
would be required to adapt our model to a new domain. 
We left SLCC out of training entirely, and trained on ChessReD \& ChessCog only: 
we used a starting model with $0\%$ board accuracy and $17.8$ wrong squares per board on SLCC.  

We then trained a low-rank adapter (LoRA) \cite{hu2021loralowrankadaptationlarge} 
on $k$ SLCC training images and measured performance on the new domain (SLCC), as well 
as the source domain (ChessReD \& ChessCog) to quantify forgetting. 
We compared four settings that differed in where the model may change and by how much:
(1) low-rank adapters on the encoder's attention and MLP projections (LoRA, rank 8, 3.1M trainable parameters), 
(2) the decoder alone (67.3M params), 
(3) the encoder alone (304.4M), and 
(4) the whole model (371.7M).  
Every mode was allotted the same $1{,}500$-step budget, a learning rate tuned per mode on target validation data, 
and the same checkpoint selection procedure, probing every 100 steps, using per-square accuracy. 
We report the mean performances over three different support-set draws.
Further protocol information is presented in the supplementary materials
(\cref{sec:supp_fewshot}, \cref{tab:fewshot}).

\begin{figure}[t]
  \centering
  \includegraphics[width=\linewidth]{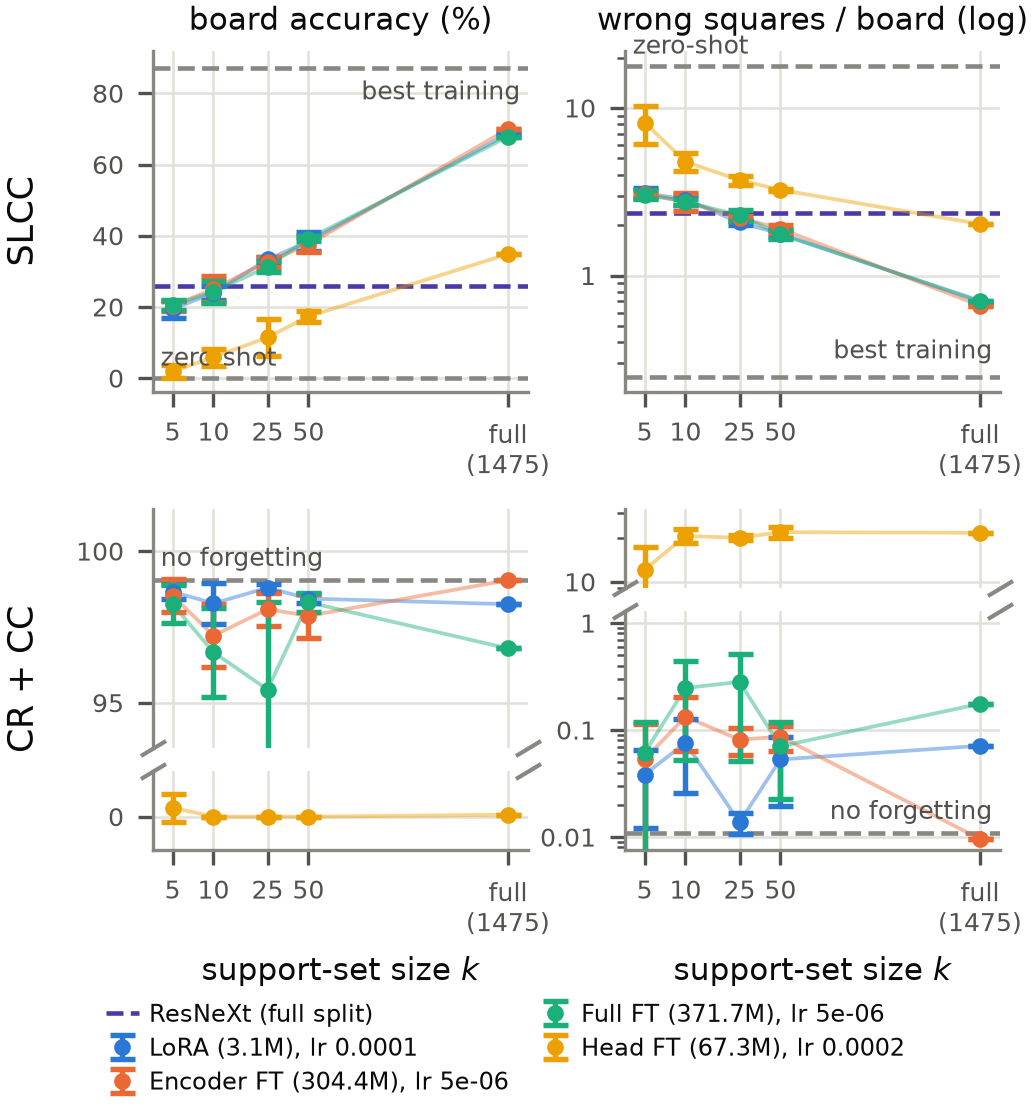}
  \caption{Few-shot adaptation of ChessQueries from ChessReD + ChessCog to the SLCC domain. \textbf{Top:} accuracy on the target (SLCC), bounded by the zero-shot floor and our best training's ceiling (87.1\% board accuracy, see \cref{tab:main}); the colored dashed line marks ChessReD's ResNeXt baseline trained on the \emph{full} SLCC split (26.0\%, \cref{tab:main}), matched by a LoRA on 10 frames. \textbf{Bottom:} forgetting is measured via the mean performance retained on the source domains (ChessReD + ChessCog). Full protocol details: \cref{sec:supp_fewshot}.}
  \label{fig:fewshot}
\end{figure}

\paragraph{Few-shot results.}
As shown in \cref{fig:fewshot}: ten training images increase 
exact board accuracy from $0\%$ to $24\%$ ($17.8 \rightarrow 2.8$ wrong squares), 
fifty images reach $38\%$ board accuracy, and the full training set $69\%$. 
This falls short of the $87\%$ that the full joint training reached (\cref{tab:main}).  

LoRA performance is similar to full fine-tuning (of either encoder or the whole model) at every $k$: 
the difference in performance lies within seed noise, 
whereas the LoRA trains only 3.1M parameters ($\sim 1\%$) instead of the 371.7M of the full model, 
resulting in a 12\,MB checkpoint in fp32.
On the other hand, decoder-only finetuning heavily underperforms on the learning task,
while catastrophically forgetting its source domain, a known pitfall to avoid~\cite{kirkpatrick2017catastrophic}.

Over the settings where the new domain is learned, we observe a negative correlation between worst-case retention 
(i.e., worst performance on ChessCog + ChessReD test sets across $k$ values) 
of the source domain and the number of parameters trained: on ChessReD board accuracy we see 
$0.965$ for LoRA, $0.948$ encoder-only, $0.878$ full fine-tuning.

We note that adaptation to the new domain primarily occurs through the encoder, not the decoder, 
which is consistent with the observations from the encoder feature pooling + linear head experiment.  
In conclusion, we find that a simple LoRA trained on ten SLCC training images is on par with the ResNeXt baseline 
\emph{trained on the full SLCC split} ($26.0\%$ board accuracy, \cref{tab:main}), and $k{=}25$ exceeds it.

\section{Limitations}
\label{sec:limitations}

One limitation encountered in this work is the SLCC dataset itself. 
For starters, its size of 2,174 images is not as large as many image datasets, 
and it could be expanded using our labeling tool; this would require 
more hours of manual labor, without any suggestion that it would significantly change our findings. 
The new SLCC dataset is hard, sometimes maybe even too hard with heavy occlusions (\cref{fig:slcc_worst}),
and in the sense of a real-world application, one could imagine that the tournament
production team would place cameras at a better angle, providing less of a challenge 
to the vision model. Our approach was to push the models to their limits, and 
that resulted in a difficult task that might not be representative of real-world use cases. 
This also highlighted the difference between $\sim 99\%$ exact-board accuracy on ChessReD / ChessCog, 
but only $\sim 87\%$ on the introduced SLCC dataset.  

ChessQueries does not handle fully side-agnostic boards, and failure cases from CVChess (in the famous Kasparov vs. Topalov game)
show that in certain positions the model can be confused about the direction in which the white and black pawns move
(see \cref{fig:cvchess_worst} in the supplementary materials).
In a practical real-world setup, it would be helpful to train the model only on images where
there is a clear signal, such as SLCC where the player with the white pieces is always on the left. 
Another interesting avenue for out-of-domain improvement would be to include an explicit modeling of the likelihood of the predicted chess positions, 
so that illegal chess positions (such as having 2 kings of the same color,~\cref{fig:cvchess_worst}) would be explicitly impossible.

Another signal that we do not exploit compared to a real-life product would be the temporal signal 
of the game of chess, where each position should only be separated from the previous one by a legal 
chess move. This is out-of-scope for our approach.  

Finally, the ablation study showed that while the query decoder approach presents multiple advantages 
in out-of-domain efficient representations and accurate square localization, 
a naive linear head nearly matches our best performance in-domain (ChessReD \& ChessCog, \cref{tab:head}). 
This indicates that a significant part of the in-domain performance improvement over the existing models 
might come from a better representation backbone, demonstrating that a ViT encoder will 
outperform a CNN-based representation such as ResNeXt or a multi-stage approach.
\section{Conclusion}
\label{sec:conclusion}

We introduced ChessQueries, a ViT encoder paired with a DETR-style decoder over 
64 learned square queries that reads a full chess board in one forward pass, 
in $19\,ms$ on a consumer GeForce RTX 4090 GPU ($\sim$590ms with a MacBook M3 Pro's MPS). 
A single architecture and training approach nearly saturates the two established public
benchmarks ($99.5\%$ exact-board accuracy on ChessReD, $98.5\%$ on ChessCog), and 
transfers zero-shot to unseen domains (CVChess). 
We also released SLCC, a 2,174-frame dataset built from
Saint Louis Chess Club professional tournament broadcast videos. 
With its occlusions, extreme viewpoints, and low-resolution boards, 
the SLCC dataset is the new frontier for chess board recognition. 
The task is genuinely challenging, and we reached $87.1\%$ exact-board accuracy, 
compared to $26\%$ for the ChessReD method. 
Our annotation pipeline is largely automatic, and the dataset could be 
expanded by applying the same method to additional chess broadcast videos. 
\anon{A model from the same family powers image import in the Lichess board
editor~\cite{lichess2026imageimport}.}{}

\clearpage
\anon{%
\section*{Acknowledgements}
I thank Luis Gon\c{c}alves, Mary Shaffer, Elena Chen, Gabriel Aubourg,
Seung-Eun Yi and Bernard Seytre for their support and help with this project.
}{}

{\small
\bibliographystyle{ieeenat_fullname}
\bibliography{references}
}

\clearpage
%
\onecolumn

\begin{center}
  {\LARGE\bf Supplementary Materials}
\end{center}
\vspace{1em}

%
%

{\centering
  \adjincludegraphics[trim={0.076\width} {0.245\height} {0.076\width}
    {0.252\height}, clip, width=\linewidth]%
    {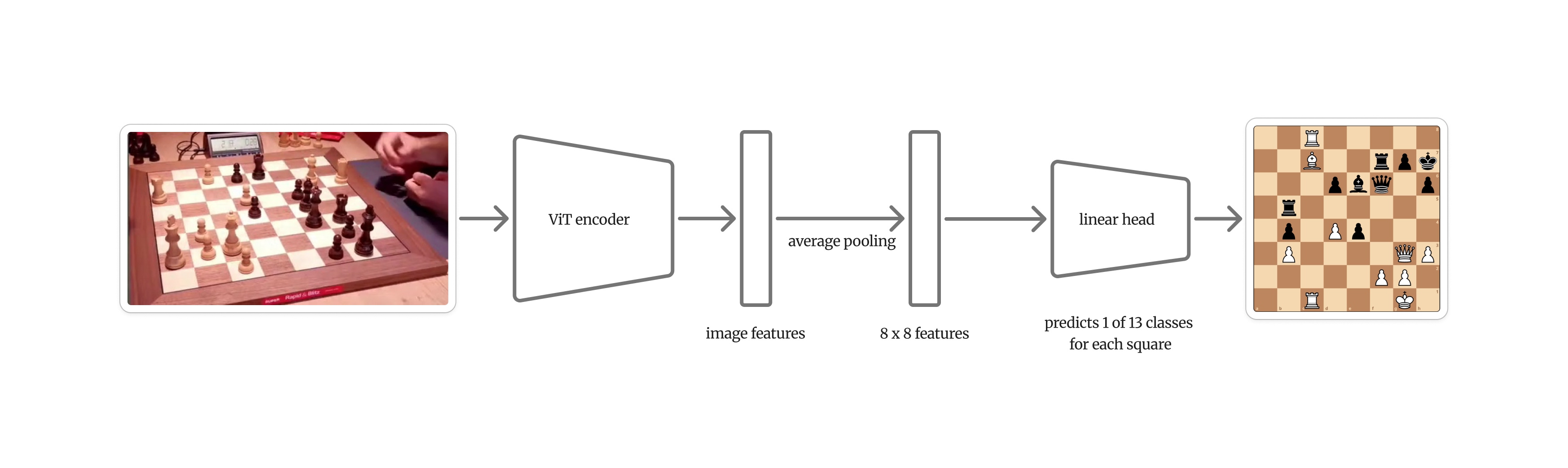}\par
}

\captionof{figure}{Architecture of the \emph{linear-head} baseline compared against in \cref{tab:head}. 
  The 64 square queries and the cross-attention decoder of \cref{fig:arch} are removed, 
  and replaced by average pooling and a simple linear head.}
\label{fig:linhead_arch}

\vspace{1em}

\appendix
\section{Worst samples per dataset}
\label{sec:supp}

\begin{figure}[b]
  \centering
  \includegraphics[width=0.75\linewidth]{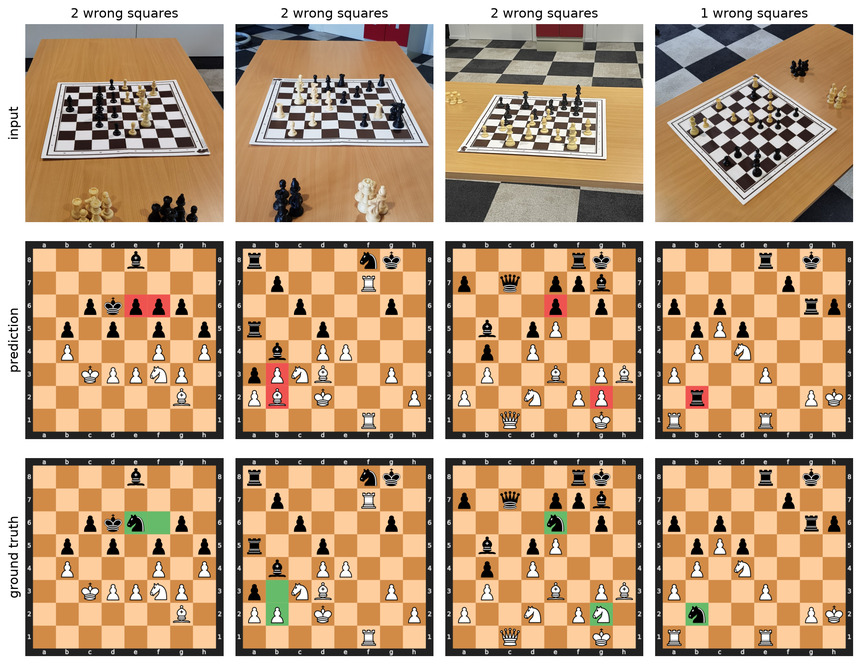}
  \caption{The four ChessReD test images where our model performs worst: at most 2 / 64 squares are wrong, mostly due to occlusion.}
  \label{fig:chessred_worst}
\end{figure}

\begin{figure}[tb]
  \centering
  \includegraphics[width=0.75\linewidth]{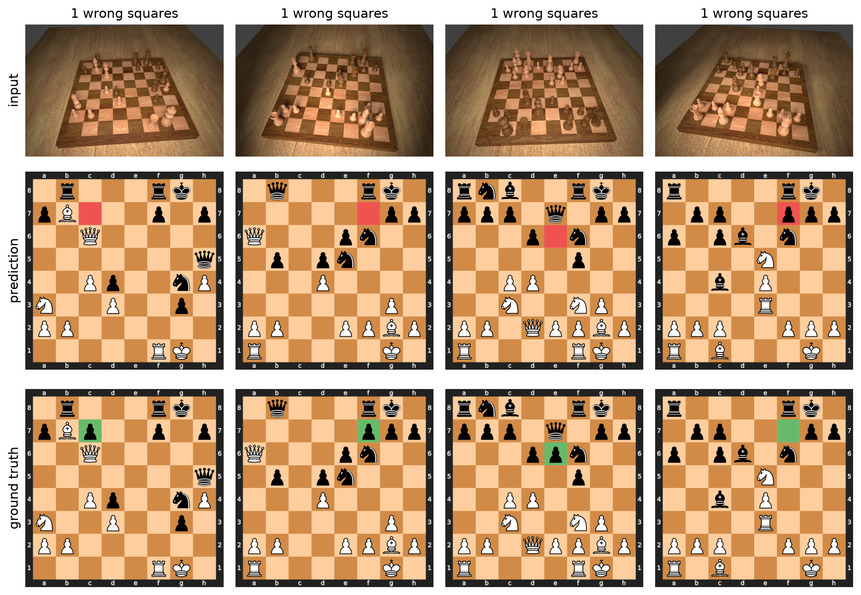}
  \caption{The four ChessCog test images where our model performs worst: a single wrong square each on this near-saturated synthetic domain, with the mistaken piece heavily occluded.}
  \label{fig:chesscog_worst}
\end{figure}

Similarly to \cref{fig:slcc_worst}, we show our model's predictions on the worst-performing samples 
of the other datasets. On the near-saturated domains (ChessReD and ChessCog) the failures are on 
at most 2 / 64 squares, under strong perspective and off-board clutter
(see \cref{fig:chessred_worst} \& \cref{fig:chesscog_worst}).  

The out-of-domain CVChess dataset shows another failure mode (see \cref{fig:cvchess_worst} and \cref{fig:cvchess_hard_angle}): 
on the unusual ending position of the famous Kasparov-Topalov game, with both kings on the first rank.
Our model fails by reading the board upside down.  
As CVChess photographs each of the 88 positions from four camera positions, we show that our model 
surprisingly performs better in viewpoints that are harder for humans. This might be due to the fact that 
the SLCC training set is exclusively made of side-views with tilted camera angles.

\begin{figure}[tb]
  \input{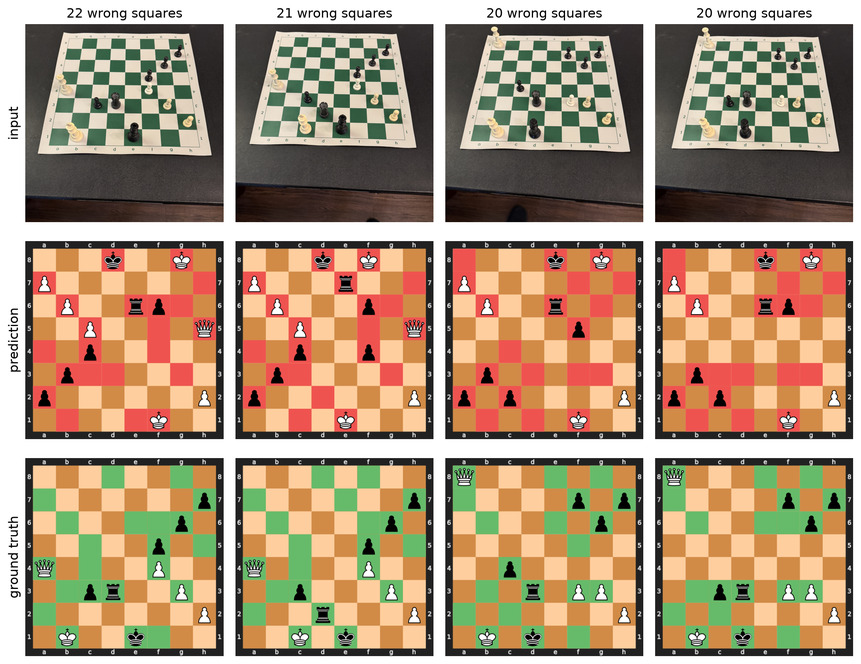}
  \par\medskip
  \input{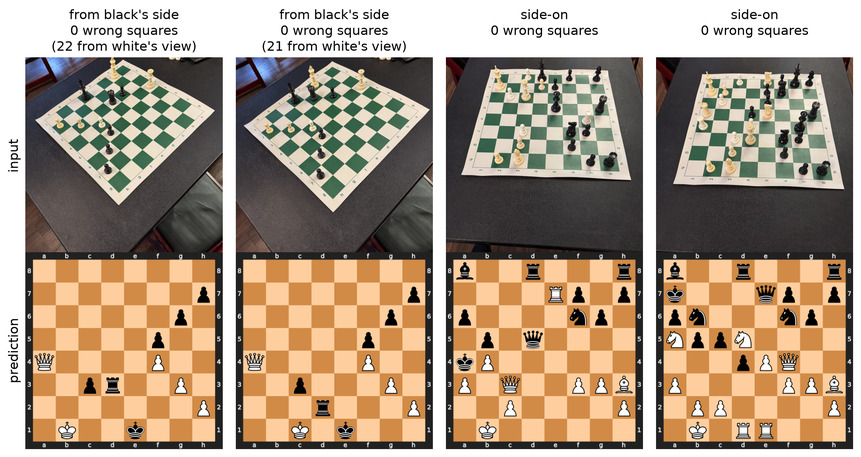}
\end{figure}

\clearpage
\section{Attention \& training the encoder}
\label{sec:supp_attention}

\begin{figure}[t]
  \centering
  \includegraphics[width=0.38\linewidth]{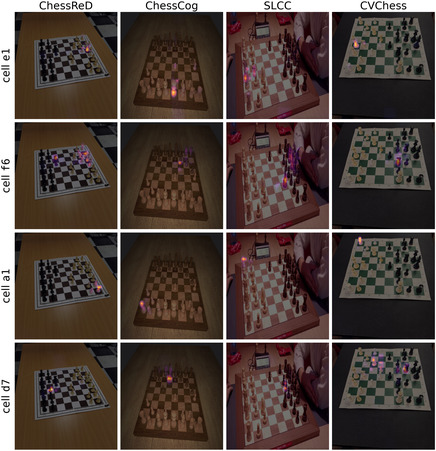}
  \caption{Encoder self-attention of the fine-tuned \emph{linear-head} model (\cref{tab:head}, top block), for the token bin each $8{\times}8$ grid cell pools over. Rows are fixed cells, columns the four domains; the shared DINOv2 global-token hotspot is removed before display. Each cell attends to a distinct on-board region that follows its square across domains: joint fine-tuning has reorganized the encoder so a readout can localize squares without a decoder.}
  \label{fig:linhead_attention}
\end{figure}

Two views complement the frozen-encoder block of \cref{tab:head}. 
The fine-tuned \emph{linear-head} model has no decoder, yet its encoder attention
is able to localize and track its square across all four domains (\cref{fig:linhead_attention}). 
That said we observe that it is less precise than the \emph{square query + decoder} approach 
as seen on \cref{fig:attention}.  

Fine-tuning therefore reorganizes the encoder into a per-square layout 
even with a simple linear head on top, and this is what makes the linear head competitive in-domain.  

\begin{figure}[b]
  \centering
  \includegraphics[width=0.70\linewidth]{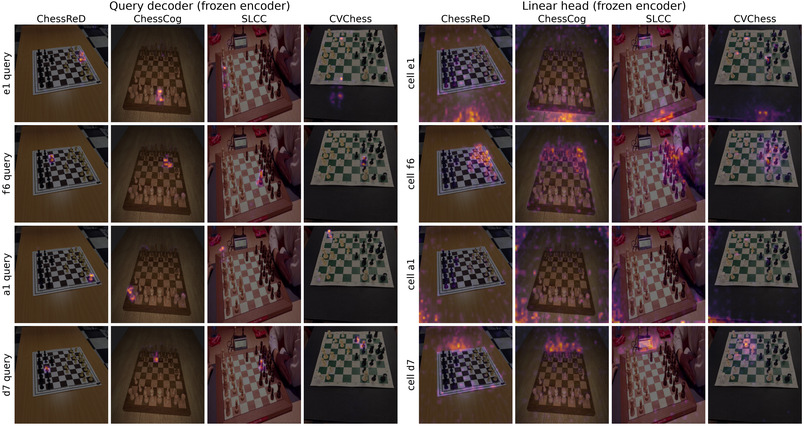}
  \caption{Attention with a \emph{frozen} encoder (\cref{tab:head}, bottom block). Left: the query decoder's cross-attention still localizes each square on frozen DINOv2 features --- the decoder computes the image-to-board correspondence itself. Right: the encoder cell attention available to the linear head is diffuse and largely off-board, with no consistent per-square localization, matching its majority-class-floor accuracy.}
  \label{fig:frozen_attention}
\end{figure}

Second, \cref{fig:frozen_attention} shows that on a \emph{frozen} encoder 
the two heads differ exactly as shown by the numbers in \cref{tab:head}: 
the query decoder's cross-attention localizes each square on frozen features (although not as well as with a trained encoder), 
while the cell attention available to the linear readout is unable to do a similar job.

\clearpage
\section{Few-shot adaptation: full results and protocol}
\label{sec:supp_fewshot}

\cref{tab:fewshot} gives the full numbers behind \cref{fig:fewshot}: exact-board
accuracy on SLCC for every set of (mode, $k$), with the trainable-parameter count
and tuned learning rate of each mode, and each mode's \emph{worst-case} source
retention over all of its runs.  

Three observations are further shown here. 
First, LoRA, encoder-only and full fine-tuning are ties at every $k$: 
all pairwise gaps are at most $2.1$ board points.  
Second, worst-case retention is negatively correlated to how many parameters are trained: 
LoRA $96.5/98.2$, encoder-only $94.8/97.7$, full fine-tune $87.8/96.5$ 
(respectively on ChessReD/ChessCog board accuracy). 
It seems that unfreezing the decoder is where retention damage comes from.  
Finally, decoder-only fine-tuning fails to learn the task appropriately, 
while collapsing on the source domain: retention per-square accuracy falls 
to $0.73$ (ChessReD) / $0.64$ (ChessCog).

\begin{table*}[t]
\centering
\adjustbox{max width=\linewidth}{%
\begin{tabular}{l@{\hskip 1.2em}cc@{\hskip 1.2em}ccccc@{\hskip 1.2em}cc}
\toprule
 &  &  & \multicolumn{5}{c}{SLCC Board\,$\uparrow$} & \multicolumn{2}{c}{Retention min.\,$\uparrow$} \\
\cmidrule(lr){4-8}\cmidrule(lr){9-10}
Mode & Trainable & LR & $k{=}5$ & $k{=}10$ & $k{=}25$ & $k{=}50$ & $k{=}\text{full}$ & ChessReD & ChessCog \\
\midrule
LoRA ($r{=}8$) & 3.1M & $1{\times}10^{-4}$ & 19.4\%$_{\pm2.4\%}$ & 23.9\%$_{\pm1.9\%}$ & 33.3\%$_{\pm0.82\%}$ & 38.4\%$_{\pm2.7\%}$ & 68.6\% & 96.5\% & 98.2\% \\
Encoder-only FT & 304.4M & $5{\times}10^{-6}$ & 20.4\%$_{\pm1.4\%}$ & 25.0\%$_{\pm3.6\%}$ & 32.7\%$_{\pm1.4\%}$ & 37.5\%$_{\pm2.1\%}$ & 70.0\% & 94.8\% & 97.7\% \\
Full FT & 371.7M & $5{\times}10^{-6}$ & 20.6\%$_{\pm1.5\%}$ & 24.2\%$_{\pm3.1\%}$ & 31.3\%$_{\pm1.6\%}$ & 39.1\%$_{\pm0.71\%}$ & 67.8\% & 87.8\% & 96.5\% \\
Decoder-only FT & 67.3M & $2{\times}10^{-4}$ & 1.9\%$_{\pm1.8\%}$ & 5.9\%$_{\pm2.4\%}$ & 11.4\%$_{\pm5.3\%}$ & 17.3\%$_{\pm1.6\%}$ & 34.9\% & 0\% & 0\% \\
\bottomrule
\end{tabular}
}
\caption{Few-shot adaptation to SLCC of the base model trained on ChessReD\,+\,ChessCog (zero-shot on SLCC: 0\% board accuracy, $17.8$ wrong squares). Each run yields mean$_{\pm\mathrm{SD}}$ over three support-set draws (except $k{=}$full is the whole $1{,}475$-frame train split, in a single run). Retention columns feature the minimum ChessReD / ChessCog board accuracy over all 13 runs of a mode.}
\label{tab:fewshot}
\end{table*}

\paragraph{Learning rate selection.}
Each mode's LR is tuned on validation data, so no tuning asymmetry favours the adapter. 
The encoder-only optimum lands at $5{\times}10^{-6}$, the same value the full fine-tune tunes to. 
Decoder-only fine-tuning has no usable window at all: six of the seven LRs probed 
end in near-total source domain collapse loss, and at the seventh, 
validation selection returns the base model.  
For the LoRA approach, a higher LR of $5{\times}10^{-4}$ earned $5$--$7$ board points 
at $k{\geq}25$ but pays for it in source domain collapse, so we report LoRA at $1{\times}10^{-4}$ throughout.

We select the final checkpoint for each run based on validation data per-square accuracy.

\paragraph{A couple comments on the approach.}
It is worth noting that all runs are based on a single starting checkpoint; 
we measure performance across three seeds with different samples of the training set, 
but we did not measure seed variance across multiple starting checkpoints.  
Compute remains modest throughout: under $8$ GPU-hours on 2 RTX~4090s 
for the whole fine-tuning experiment including the LR sweeps, and the final best LoRA weights 
take up $12$\,MB, to be attached to a $1.5$\,GB checkpoint.

\clearpage
\section{SLCC annotation and reconstruction pipeline}
\label{sec:supp_slcc_annotation}
\begin{figure}[b]
  \centering
  \makebox[\linewidth][c]{%
    \includegraphics[width=1.10\linewidth]{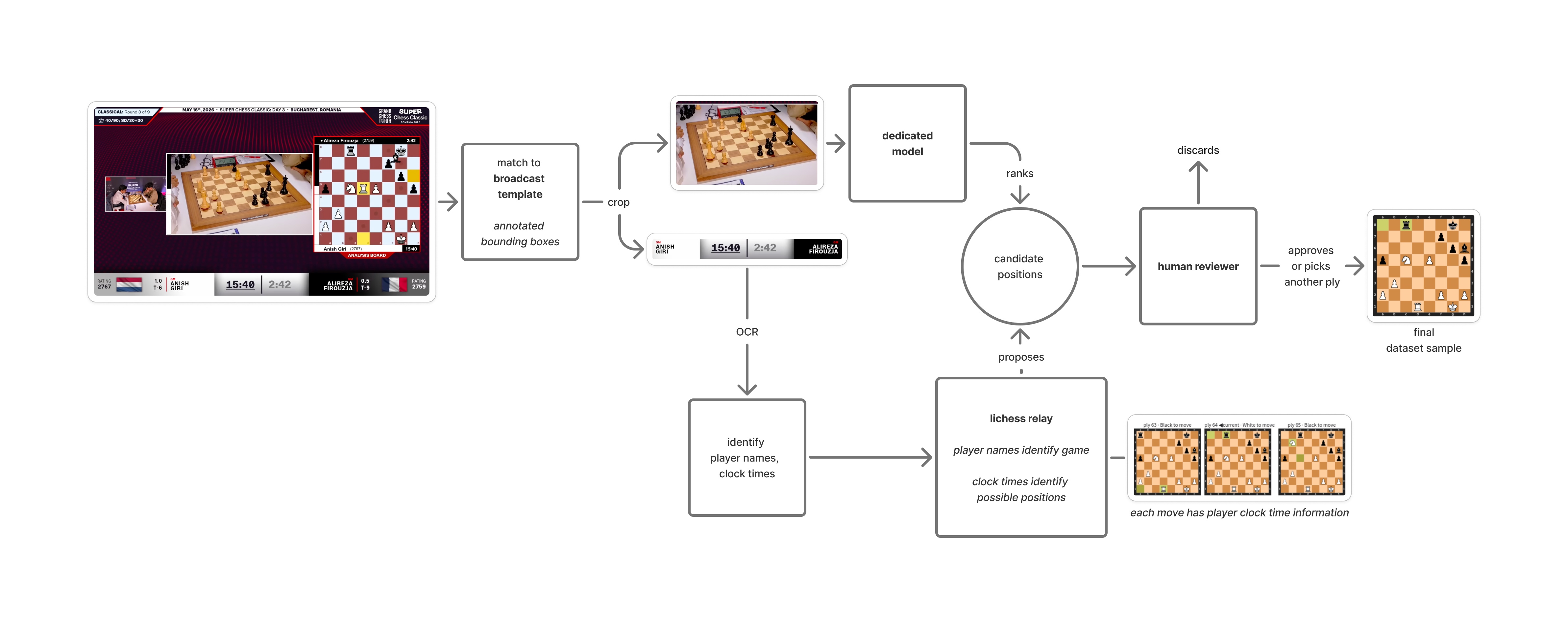}%
  }
  \caption{Overview of the semi-automated SLCC annotation and reconstruction pipeline.
  Lichess~\cite{lichess} is an online platform that provides chess features such as online play, 
  analysis, or information on current and past tournaments,
  including the results, the moves played, remaining time for each player after each move, etc.}
  \label{fig:slcc_annotation_pipeline_diagram}
\end{figure}

\begin{figure}[tb]
  \centering
  \includegraphics[width=0.55\linewidth]{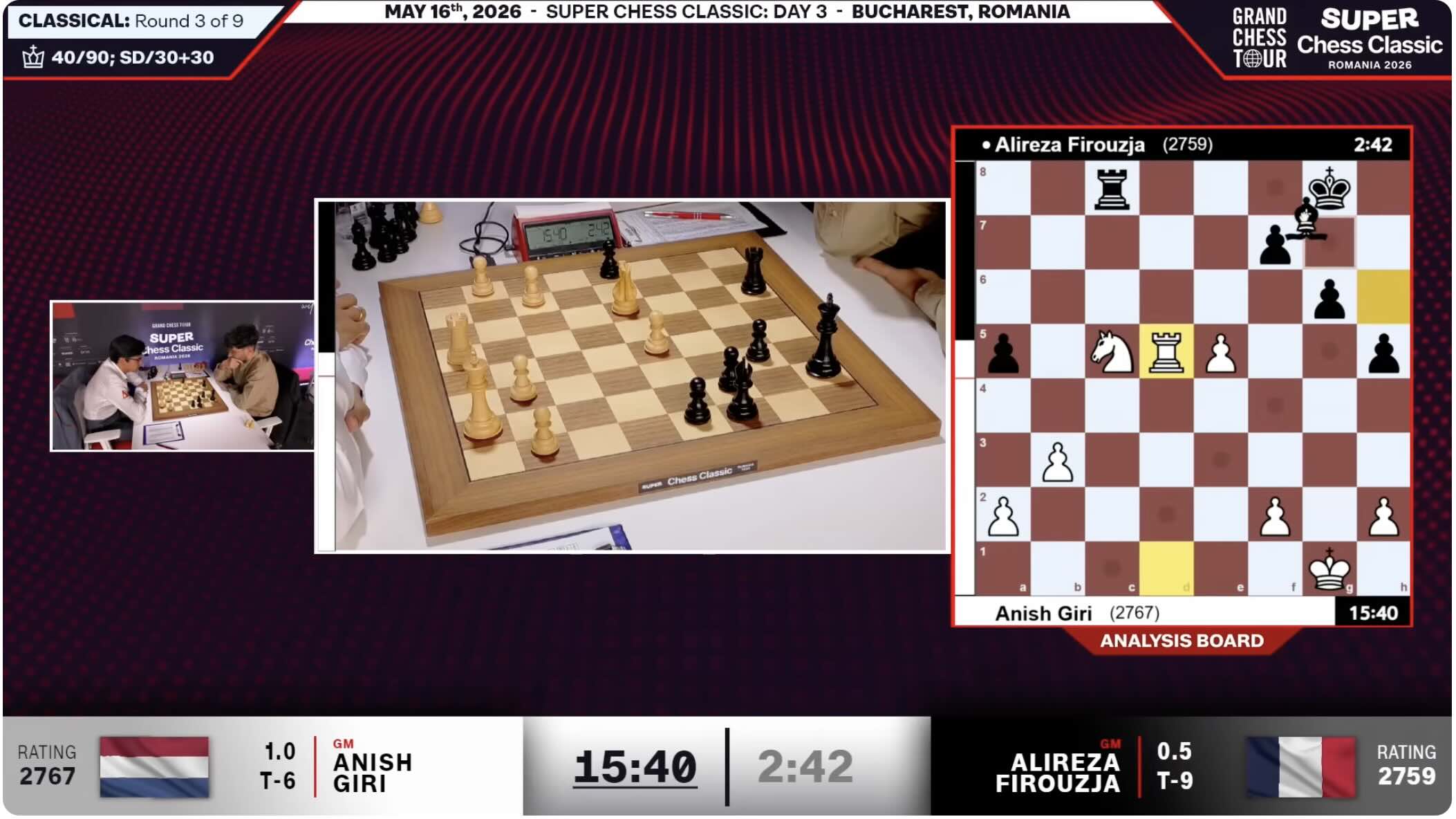}
  \par\vspace{2pt}
  {\color{black!35}\rule{0.78\linewidth}{0.4pt}}
  \par\vspace{2pt}
  \includegraphics[width=0.73\linewidth]{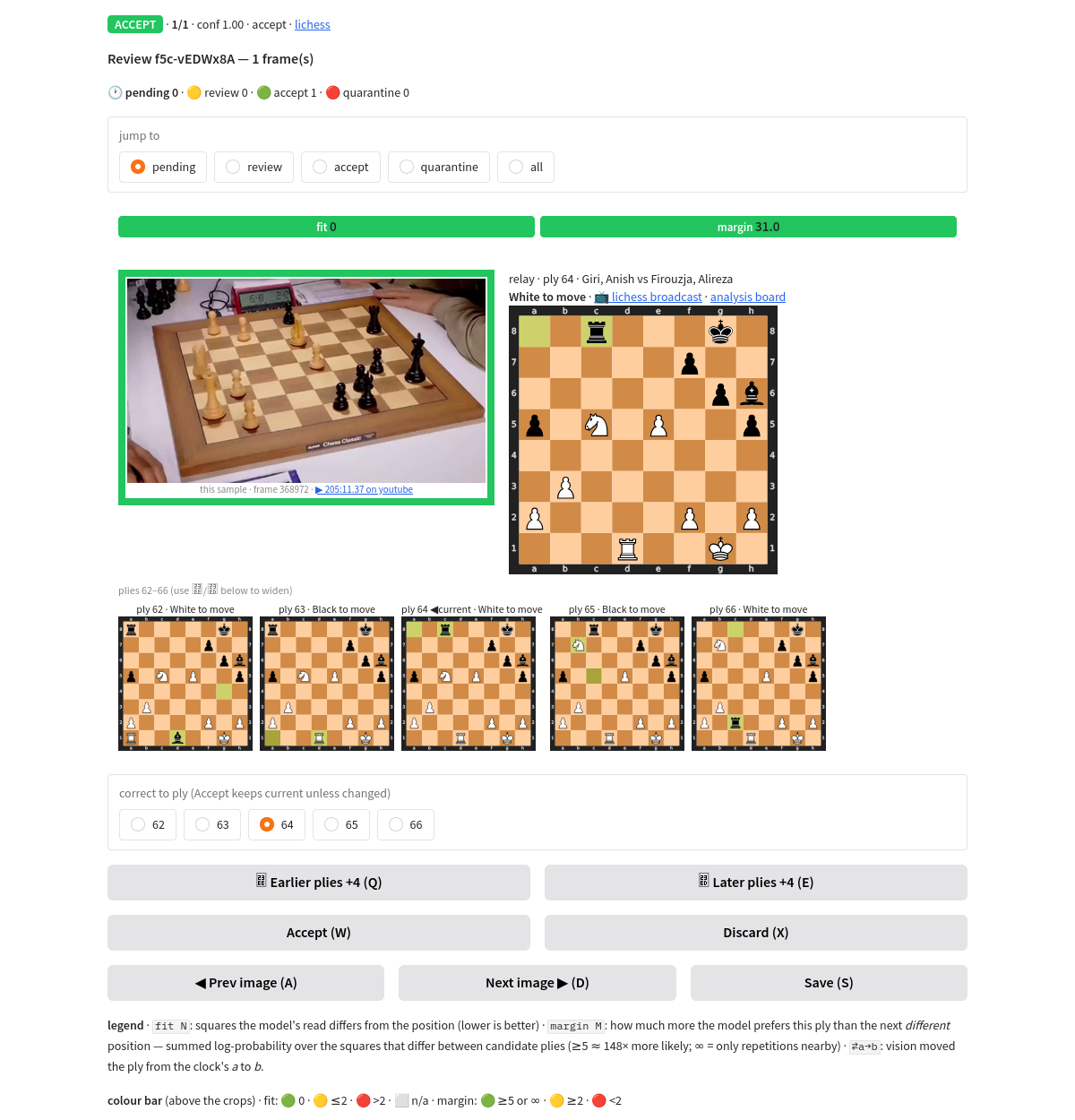}
  \caption{\textbf{Top:} Example of a SLCC broadcast layout~\cite{stlcc,GCT}. 
  We crop out the physical-board view as the main model input, 
  while OCR reads (1) the player names and (2) their remaining clock
  times (here: $15\,\mathrm{min}\,40\,\mathrm{s}$ vs.\ $2\,\mathrm{min}\,42\,\mathrm{s}$).
  We deliberately ignore the analysis board on the right because it often shows
  commentary positions rather than the live game. \textbf{Bottom:}
  The annotation interface to review the candidate annotation extracted from the broadcast. 
  Here both the OCR $\rightarrow$ Lichess relay pipeline's candidate and the LoRA model both select ply 64; 
  the model agrees with the relay position on all 64 squares (\texttt{fit 0}), 
  with a predicted log-probability margin of 31.0 over the next closest candidate. 
  The human reviewer must now inspect the neighboring plies, and decide: accept, 
  correct which ply of the game fits the image, or discard the sample.}
  \label{fig:slcc_annotation_pipeline}
\end{figure}

We summarize the semi-automated process in \cref{fig:slcc_annotation_pipeline_diagram}.
The YouTube broadcast (\cref{fig:slcc_annotation_pipeline}) comes in a layout 
that needs to be parsed into a proper chess board image and associated chess position.
We manually annotate templates, which include bounding box information, to locate production elements within a given frame: 
(1) the main board image; (2) the players' names; (3) their remaining clock times.

Through the YouTube video id, each frame is mapped to its tournament round, 
and a given pair of opponents plays only one game in each selected video, thus we obtain the correspondence between frame and chess game.
Additionally, through the Lichess relay~\cite{lichess} we know the move sequence and clock state after 
each ply, providing candidate positions.
Unfortunately, the clocks do not necessarily identify a unique ply: 
with increments, players gain time after making a move, so the same pair of 
displayed times may correspond to multiple positions.
We also observed occasional delays in the broadcast between the 
camera feed and the clock overlay, creating further ambiguity.
We therefore retain a list of candidate relay positions rather than a single identified position.

We resolved this ambiguity with a ChessQueries-based model.
We first trained a version of ChessQueries (V0) only on ChessReD and ChessCog, 
and manually annotated the first 20 SLCC images without any model assistance.
We then trained a LoRA of V0 on those 20 images, which were later assigned to the SLCC training split.
From there on, the model is used to rank the candidate positions, and shows the annotator the most likely one.
Note that the model does \emph{not} generate the final FEN annotation, and that this model was then discarded; its weights are unrelated 
to those of the final ChessQueries model presented in the main section of the paper.

Finally, a human eye reviews the proposed match in the interface shown in~\cref{fig:slcc_annotation_pipeline}.
The reviewer can check the information, accept the candidate, manually choose a different ply, or discard it.
Every retained sample is human-verified, and we observed no incorrect match when the OCR-derived position 
and visual ranker agreed (on \texttt{fit} and \texttt{margin} criteria shown in the figure).

Using a model to help us annotate allowed us to go much faster, and we needed to make very few edits.
It would have been unfeasible to review and accept 2,174 images without model assistance.


\end{document}